\documentclass[runningheads]{llncs}
\usepackage[T1]{fontenc}
\usepackage{graphicx}
\usepackage{tcolorbox}
\usepackage{subcaption} 
\usepackage{color}
\usepackage{amsmath}
\usepackage{amssymb}
\usepackage{subcaption} 

\begin{document}
\title{Comparative reversal learning reveals rigid adaptation in LLMs under non-stationary uncertainty}
\titlerunning{Rigid adaptation in LLM reversal learning}

\author{
Haomiaomiao Wang\inst{1}\orcidID{0009-0005-5961-1847} 
\and
Tomás E Ward\inst{1,2}\orcidID{0000-0002-6173-6607}
\and
Lili Zhang\inst{1,2}\orcidID{0000-0002-2203-2949}
}

\authorrunning{H. Wang et al.}
\institute{Insight Research Ireland Centre for Data Analytics, Ireland\and School of Computing, Dublin City University, Ireland }
\maketitle              
\begin{abstract}
Non-stationary environments require agents to revise previously learned action values when contingencies change. We treat large language models (LLMs) as sequential decision policies in a two-option probabilistic reversal-learning task with three latent states and switch events triggered by either a performance criterion or timeout. We compare a deterministic fixed transition cycle to a stochastic random schedule that increases volatility, and evaluate DeepSeek-V3.2, Gemini-3, and GPT-5.2, with human data as a behavioural reference. Across models, win-stay was near ceiling while lose-shift was markedly attenuated, revealing asymmetric use of positive versus negative evidence. DeepSeek-V3.2 showed extreme perseveration after reversals and weak acquisition, whereas Gemini-3 and GPT-5.2 adapted more rapidly but still remained less loss-sensitive than humans. Random transitions amplified reversal-specific persistence across LLMs yet did not uniformly reduce total wins, demonstrating that high aggregate payoff can coexist with rigid adaptation. Hierarchical reinforcement-learning (RL) fits indicate dissociable mechanisms: rigidity can arise from weak loss learning, inflated policy determinism, or value polarisation via counterfactual suppression. These results motivate reversal-sensitive diagnostics and volatility-aware models for evaluating LLMs under non-stationary uncertainty.
\keywords{Reversal learning \and Non-stationary uncertainty \and LLMs}
\end{abstract}

\section{Introduction}
Non-stationary uncertainty is ubiquitous in real decision settings. Reward structures drift or reverse after policy changes or market shocks \cite{xieDeepReinforcementLearning2020,cheungNonstationaryReinforcementLearning2023}. When an agent fails to adapt, the cost is not merely suboptimal reward. It can be financial loss or operational risk \cite{zhuangConsequencesMisalignedAI2021}. This concern is becoming more salient as LLMs move from passive tools to delegated decision partners in everyday workflows \cite{passiAgenticAIHas2025}.

A growing literature therefore treats LLMs as sequential decision-makers in bandits and RL tasks\cite{wang2026rigidityllmbanditsimplications,zhangAdversarialTestingLLMs2025,zhangLandscapeAgenticReinforcement2025,nieEVOLvEEvaluatingOptimizing2024}. These studies show that LLMs can update their choices based on feedback. But their behaviour often deviates from statistically optimal or theoretically expected forms of adaptation. Computational cognitive modelling has been particularly informative in this context \cite{wang2026rigidityllmbanditsimplications,schaafTestRetestReliability2023,vangeenHierarchicalBayesianModels2021}. Because it maps observed choice sequences onto interpretable parameters, such as learning rates and choice determinism. Our prior results in stationary bandit tasks illustrate the pattern. LLMs achieve high average performance while still exhibiting rigidity at the level of underlying decision processes.

However, stationary bandit tasks characterise uncertainty only as randomness around fixed reward distributions \cite{wang2026rigidityllmbanditsimplications}. They do not examine whether an agent can revise previously learned beliefs when the environment itself changes. This limitation highlights the distinction between changing and non-stationary environments. Changing is an informal term that can refer to any variation in inputs. Non-stationary, by contrast, has a precise technical meaning. The data-generating process evolves over time, so action-outcome relationships that were previously valid may no longer hold \cite{xieDeepReinforcementLearning2020}. Strong performance in stationary tasks can conceal brittle strategies that fail once reinforcement contingencies reverse.

To address this gap, we use reversal learning to study non-stationary uncertainty in a controlled two-option task, a minimal setting that isolates adaptive flexibility and enables interpretable analysis. Such tasks are typically used as mechanistic diagnostics, with generalisation to adversarial attacks or shifting market conditions \cite{leibo2021scalable}. We examine two reversal conditions. In the fixed condition, reversals occur at predetermined trial points, creating segments with stable reward structure. In the random condition, reversals occur unpredictably by introducing persistent volatility and uncertainty about when changes will happen. Multiple LLMs are evaluated with human behaviour as a reference where appropriate \cite{schaafTestRetestReliability2023}. We quantify reversal-specific behavioural costs first. Then we fit hierarchical RL models to distinguish between competing mechanisms.

\section{Methods}

\subsection{Models and decoding settings}
\label{sec:models}

We evaluated three representative LLMs as sequential decision policies in a two-option reversal-learning task. All models were queried using fixed decoding settings in the main experiments (Table~\ref{tab:llm_api_decoding}). In prior work, we found that varying temperature and top-$p$ does not qualitatively change learning patterns in minimal bandit tasks, but mainly modulates surface-level output variability \cite{wang2026rigidityllmbanditsimplications}. We confirmed this in pilot bandit runs by comparing the default configuration with lower-temperature, benchmark-oriented settings reported in competitive evaluations. These alternative settings did not improve and in several cases reduced the optimal-choice rate. We therefore retained the default decoding configuration for all models to maximise stability and comparability across model families.

\begin{table}[h]
\centering
\caption{Evaluated LLMs and decoding settings used for data generation.}
\label{tab:llm_api_decoding}
\begin{tabular}{|l|l|l|l|l|}
\hline
\textbf{Provider} & \textbf{Model} & \textbf{API model ID} & Temperature & top-$p$ \\
\hline
DeepSeek & DeepSeek-V3.2 & \texttt{deepseek-chat} & 1.0 & 1.0 \\
Google  & Gemini-3-Flash & \texttt{gemini-3-flash-preview} & 1.0 & 1.0 \\
OpenAI  & GPT-5.2 & \texttt{gpt-5.2} & 1.0 & 1.0 \\
\hline
\end{tabular}
\end{table}

\subsection{Reversal-learning task and experimental design}
\label{sec:reversal_design}

For each model, we ran $N=200$ independent runs per schedule condition, with $T=250$ trials per run. On each trial $t$, the model selected one of two options and received binary feedback, coded as $+100$ (win) or $-100$ (loss).
Outcome probabilities were governed by a latent task state that was not revealed to LLMs.

\paragraph{Latent states.}
Let $s_m \in \{0,1,2\}$ denote the latent state in segment $m$ with
\[
s = 0:\ [0.80,0.20], \qquad
s = 1:\ [0.50,0.50], \qquad
s = 2:\ [0.20,0.80],
\]
where each pair denotes $[P(\mathrm{win}\mid E),\, P(\mathrm{win}\mid V)]$.
Within each state, outcomes are sampled independently according to the Bernoulli probability of a win for the chosen option. All runs start in state $s_0 = 0$.

\paragraph{Switch trigger.}
The environment remained in the current state until a switch was triggered by either a performance criterion or a timeout.
We tracked a rolling window of the last $10$ choices within the current state.
If at least $7$ of these choices matched the state-defined target option, a switch was triggered.
If the state persisted for $16$ trials without meeting this criterion, a switch was triggered automatically.
After each switch, the within-state choice history was reset.

The target option was defined as the option with higher reward probability in the current state.
In the tie state $[0.50,0.50]$, the target used for the criterion rule was defined as the option that had been less rewarding in the immediately preceding non-tie state.
This convention was used only to evaluate the performance criterion and did not expose reward probabilities to the model.

\paragraph{Fixed versus random schedules.}
Both conditions used the same latent state space and the same switch-trigger rule.
They differed only in how the next state was selected after a switch.
Here, ``fixed'' refers to a deterministic ordering of states rather than fixed-length blocks.

\begin{enumerate}
    \item \textbf{Fixed schedule (deterministic cycle).}
    After each switch, the next state followed the cycle
    \[
    0 \rightarrow 1 \rightarrow 2 \rightarrow 0 \rightarrow \cdots,
    \]
    equivalently $s_{m+1} = (s_m + 1)\bmod 3$.

    \item \textbf{Random schedule (stochastic next state).}
    After each switch, the next state was sampled uniformly from the two alternative states:
    \[
    s_{m+1} \sim \mathrm{Unif}\bigl(\{0,1,2\}\setminus\{s_m\}\bigr).
    \]
    With three states, this implies $P(s_{m+1}=j\mid s_m=i)=1/2$ for each $j\neq i$.
\end{enumerate}

\subsection{Prompting protocol and response validation}
\label{sec:prompting}

On each trial, the model received a fixed-format prompt comprising (i) a system instruction specifying the goal and output constraint, and (ii) a user message reporting the current trial index and a running history of prior choices and outcomes.
The model was instructed to output exactly one uppercase character corresponding to its choice.
In the main experiments, the action alphabet was $\{\texttt{E}, \texttt{V}\}$.
Responses were parsed strictly: a response was valid only if it was exactly \texttt{E} or \texttt{V} after trimming whitespace.

If the output was invalid, we retried the same trial with an explicit corrective instruction, up to a maximum of 5 retries.
If the model remained non-compliant after the retry budget, the run was terminated and marked as incomplete.
Invalid-output rates were logged and analysed separately.

\begin{tcolorbox}[title=Example prompt, colback=gray!5!white, colframe=black!75!black]
\textbf{System:}
You are a space explorer choosing between two planets, \texttt{E} and \texttt{V}.
On every trial, the planet you choose results in either a gain of 100 gold coins (\(+100\))
or a loss of 100 gold coins (\(-100\)).
Throughout the mission, it may change multiple times which planet is more likely to yield \(+100\)
and which is more likely to yield \(-100\).
Feedback is probabilistic: even if you choose the planet that is more likely to yield \(+100\),
you may still receive \(-100\).
Your goal is to maximise your total number of gold coins over exactly 250 trials.
Do not output any other words, punctuation, or explanations.
You must respond with exactly one uppercase character: \texttt{E} or \texttt{V}.

\medskip
\textbf{Prompt:}
You have completed 3 of 250 trials.\\
History:\\
\quad - Trial 1: choice=\texttt{E}, outcome=\(+100\)\\
\quad - Trial 2: choice=\texttt{E}, outcome=\(-100\)\\
\quad - Trial 3: choice=\texttt{V}, outcome=\(-100\)\\
Choose for Trial 4.\\
Answer:
\end{tcolorbox}

\subsection{Prompt-format controls and label selection}
\label{sec:prompt_controls}

To test whether lexical or positional biases could confound learning, we conducted two prompt-format controls:
\begin{enumerate}
    \item \textbf{Order control:} we reversed the presentation order of the two options, ``E and V'' vs.\ ``V and E''.
    \item \textbf{Label control:} we replaced the action labels (\texttt{X}/\texttt{Y} or \texttt{W}/\texttt{L}) in place of \texttt{E}/\texttt{V}.
\end{enumerate}

For each variant, we evaluated first-trial choice bias and the learning curve for the optimal option, and summarised late-trial choice proportions over the final 50 trials. Across variants, models showed similar learning curves and convergence to the higher-reward option, indicating that behaviour was not driven by label or order effects. We selected \texttt{E}/\texttt{V} for the main experiments because it was less sensitive to label order while preserving convergence, reducing the risk that early prompt artefacts dominate long-run behaviour.

\subsection{Behavioural measures}
\label{sec:behavior}

Behavioural analyses were performed at the run level and aggregated within each model $\times$ schedule cell, treating each run as a simulated participant. Trials are indexed by $t \in \{1,\dots,T\}$ with actions $a_t \in \{\texttt{E},\texttt{V}\}$; the unchosen action is denoted $a'_t \neq a_t$. Outcomes are coded as a win indicator $w_t \in \{0,1\}$ ($w_t=1$ for $+100$, $w_t=0$ for $-100$); equivalently, we use the rescaled feedback $r_t = 2w_t-1 \in \{+1,-1\}$. The latent task state at trial $t$ is $s_t \in \{0,1,2\}$ and is piecewise constant within segments between true task switches (Sec.~\ref{sec:reversal_design}).

To avoid mixing behaviour across contingency changes, sequential statistics were computed only for adjacent trial pairs that remained in the same segment, i.e.\ $s_t=s_{t-1}$. Define $\mathbb{I}[\cdot]$ as the indicator function. Lose--shift was defined as the conditional probability of switching after a loss:
\[
P(\mathrm{shift}\mid \mathrm{loss}) \;=\;
\frac{\sum_{t=2}^{T} \mathbb{I}[a_t\neq a_{t-1}]\,\mathbb{I}[w_{t-1}=0]\,\mathbb{I}[s_t=s_{t-1}]}
{\sum_{t=2}^{T} \mathbb{I}[w_{t-1}=0]\,\mathbb{I}[s_t=s_{t-1}]} .
\]
(Analogously, win-stay uses $\mathbb{I}[a_t=a_{t-1}]$ and $\mathbb{I}[w_{t-1}=1]$.)

\paragraph{Reversal-sensitive measures.}
Segments were identified directly from the task metadata (state entry indices or state identifiers), ensuring that reversal boundaries corresponded to true contingency changes. Let segment $m$ begin at the first post-switch trial $t_r^{(m)}$ and end at $t_{\mathrm{end}}^{(m)}$, with latent state $s^{(m)} \in \{0,1,2\}$. Reversal-sensitive measures were evaluated only for segments with a unique optimal action (i.e.\ $s^{(m)}\in\{0,2\}$). Define the optimal action under state $s$ as
\[
a^\star(s) \;=\; \arg\max_{a \in \{\texttt{E},\texttt{V}\}} P(\mathrm{win}\mid a,s),
\]
so that $a^\star(0)=\texttt{E}$ and $a^\star(2)=\texttt{V}$.

\paragraph{Perseveration length.}
For each eligible segment $m$, let $a^\star_{\mathrm{prev}}$ denote the optimal action in the most recent preceding eligible (non-tie) segment. Perseveration length was defined as the number of consecutive trials after the switch on which the agent continued to select $a^\star_{\mathrm{prev}}$:
\[
L^{(m)} \;=\; \sum_{k \ge 0} \mathbb{I}\!\left[a_{t_r^{(m)}+k}=a^\star_{\mathrm{prev}}\right],
\]
where counting stops at the first deviation from the previous optimal action.

\paragraph{Post-reversal regret.}
We computed trial-wise regret from the known reward probabilities,
\[
\mathrm{Regret}_t \;=\; \max_{a}P(\mathrm{win}\mid a,s_t)\;-\;P(\mathrm{win}\mid a_t,s_t),
\]
and summed regret within a fixed post-reversal window of $W=20$ trials (or until the segment ended):
\[
C^{(m)} \;=\; \sum_{t=t_r^{(m)}}^{\min(t_r^{(m)}+W-1,\;t_{\mathrm{end}}^{(m)})} \mathrm{Regret}_t .
\]
For interpretability, regret was additionally expressed in expected coins by multiplying by 100. For each run, reversal-sensitive measures ($L^{(m)}$ and $C^{(m)}$) were computed per eligible segment and averaged within run.

\paragraph{Overall task performance.}
Total wins were defined as $\mathrm{Wins}=\sum_{t=1}^{T} w_t$. Runs that terminated early due to persistent invalid outputs contributed outcomes up to termination and were excluded from trial-based summaries beyond that point.

\subsection{Computational models}
\label{sec:comp_models}

We fitted hierarchical RL models to the trial-level choice and outcome sequences $\{a_t,r_t\}_{t=1}^{T}$ (notation as in Sec.~\ref{sec:behavior}) to obtain mechanistic parameters governing (i) value updating and (ii) choice stochasticity.

\paragraph{Value and choice rule.}
Let $Q_{a,t}$ denote the value of action $a\in\{\texttt{E},\texttt{V}\}$ at the start of trial $t$, with $Q_{\texttt{E},1}=Q_{\texttt{V},1}=0$. Choices follow a logistic softmax on the value difference,
\[
P(a_t=\texttt{E})=\sigma\!\left(\beta\,(Q_{\texttt{E},t}-Q_{\texttt{V},t})\right),
\qquad
\sigma(x)=\frac{1}{1+\exp(-x)},
\]
where $\beta>0$ is the inverse temperature.

\paragraph{Dual learning-rate RL model (Dual RL).}
After observing outcome $r_t$ following choice $a_t$, the prediction error is $\delta_t=r_t-Q_{a_t,t}$. The chosen value updates with separate learning rates for gains and losses,
\[
Q_{a_t,t+1} =
Q_{a_t,t} +
\begin{cases}
\eta_{\mathrm{pos}}\,\delta_t, & r_t = +1,\\
\eta_{\mathrm{neg}}\,\delta_t, & r_t = -1,
\end{cases}
\qquad
Q_{a'_t,t+1}=Q_{a'_t,t},
\]
where $\eta_{\mathrm{pos}},\eta_{\mathrm{neg}}\in(0,1)$.

\paragraph{Dual RL with partial double updating (Dual RL-$\kappa$DU).}
We also fitted a partial double-updating extension in which the unchosen action receives a counterfactual outcome of $-r_t$ (reflecting anticorrelated options; \cite{schaafTestRetestReliability2023}). The unchosen counterfactual prediction error is
\[
\delta^{\mathrm{unch}}_t = (-r_t) - Q_{a'_t,t},
\]
and the unchosen value additionally updates with the same valence-dependent learning rate, scaled by $\kappa \in (0,1)$:
\[
Q_{a'_t,t+1}
=
Q_{a'_t,t}
+
\kappa
\begin{cases}
\eta_{\mathrm{pos}}\,\delta^{\mathrm{unch}}_t, & r_t=+1,\\
\eta_{\mathrm{neg}}\,\delta^{\mathrm{unch}}_t, & r_t=-1.
\end{cases}
\]
Thus, $\kappa=0$ recovers Dual RL, while larger $\kappa$ increases counterfactual updating.

\paragraph{Why fitted $\eta_{\mathrm{pos}}$ can become large in reversal tasks.}
Because choice depends on the scaled value difference $\beta\,\Delta Q_t$ with $\Delta Q_t=Q_{\texttt{E},t}-Q_{\texttt{V},t}$, strong win-stay can be captured either by increasing $\beta$ or by increasing how quickly $\Delta Q_t$ grows after wins (larger effective gain updating). Under $\kappa$DU and neutral initialization, a single win ($r_t=+1$) yields
\[
Q_{a_t,t+1}=\eta_{\mathrm{pos}},\qquad
Q_{a'_t,t+1}=-\kappa\,\eta_{\mathrm{pos}}
\quad\Rightarrow\quad
|\Delta Q_{t+1}|=\eta_{\mathrm{pos}}(1+\kappa),
\]
so that repeat-choice probability after a win is approximately
\[
P(\text{repeat}\mid r_t=+1)\approx \sigma\!\left(\beta\,\eta_{\mathrm{pos}}(1+\kappa)\right).
\]
Accordingly, if empirical win-stay is high while $\beta$ and/or $\kappa$ are moderate, posterior mass can shift toward larger $\eta_{\mathrm{pos}}$ to match the observed persistence. For hierarchical RL fitting, incomplete runs were included up to the last valid trial, and no missing responses were imputed.

\subsection{Hierarchical estimation and comparison}
We fit hierarchical Bayesian versions of the RL models in JAGS, estimating run-level
$\eta_{\mathrm{pos}}, \eta_{\mathrm{neg}}, \beta$ (and $\kappa$ for $\kappa$DU) with weakly informative bounded priors.
Model adequacy was evaluated with posterior predictive checks; Dual RL and Dual RL-$\kappa$DU were compared using DIC.
Implementation details are omitted for space \cite{schaafTestRetestReliability2023}.

\section{Results}

\subsection{Behavioural metrics of rigid adaptation}
\label{sec:results_behaviour}

\begin{figure}[h]
    \centering
    \includegraphics[width=\linewidth]{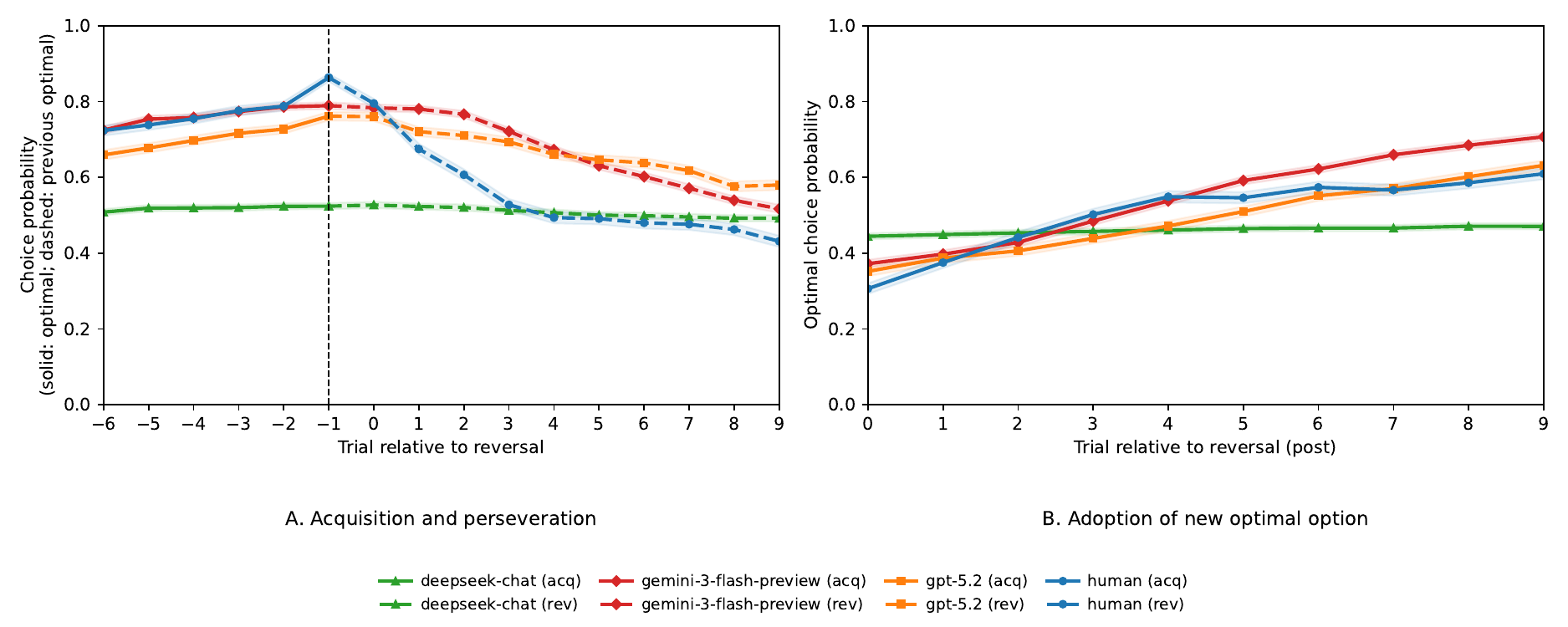}
    \caption{Random schedule: post-reversal adoption of the target option and event-aligned choice proportions around reversals.}
    \label{fig:behavioural}
\end{figure}

We first examined how rapidly agents shifted toward the newly advantageous option following reversals under the random schedule. Choice proportions were aligned to the reversal boundary to quantify pre- and post-reversal situations. Humans, Gemini-3, and GPT-5.2 exhibited a clear discontinuity at the boundary, consistent with abrupt updating when contingencies changed (Fig.~\ref{fig:behavioural}A). DeepSeek-V3.2 showed weaker acquisition prior to reversals and a markedly attenuated reorientation afterward, indicating reduced sensitivity to contingency change. Behaviour was robust to action label and order manipulations, with similar asymptotic performance and reversal-sensitive metrics across variants. However, robustness to broader prompt wording or narrative framing remains for future work. Invalid outputs occasionally required retries; however, all trials yielded valid responses within the retry limit, and no runs were terminated.

Across the first ten post-reversal trials (Fig.~\ref{fig:behavioural}B), Gemini-3 showed the clearest adaptation: target-option probability increased monotonically. GPT-5.2 exhibited a similar but slower increase. In contrast, DeepSeek-V3.2 remained close to indifference, reflecting persistent choice inertia despite continued feedback.

To characterise reversal behaviour more formally, we computed complementary metrics capturing switching, persistence, and reward accumulation (Table~\ref{tab:beh_metrics}). Three robust patterns emerged.

\paragraph{LLMs exhibited near-ceiling win–stay but attenuated lose–shift.} All LLMs showed extremely high win–stay probabilities ($>0.93$), indicating strong reinforcement of rewarded actions. However, lose–shift probabilities diverged sharply. DeepSeek-V3.2 exhibited minimal switching after losses. Gemini-3 and GPT-5.2 showed higher lose–shift, but remained well below the human reference. Thus, LLMs preferentially reinforced gains but under-reacted to losses relative to humans.

\paragraph{Perseveration differed sharply across agents.}
DeepSeek-V3.2 showed substantially longer perseveration following true reversals than Gemini-3 and GPT-5.2, while humans exhibited minimal persistence. Switch latency showed the same ordering: humans typically selected the new optimal action within one trial, Gemini-3 and GPT-5.2 within 3–7 trials, and DeepSeek-V3.2 substantially later under the fixed schedule. Thus, DeepSeek's rigidity reflects sustained continuation of the previously optimal action rather than merely delayed initial switching.

\paragraph{Random schedules increased persistence but reduced post-reversal regret.}
Across LLMs, random transitions increased perseveration relative to fixed schedules. However, post-reversal regret decreased under randomness for all three models. For example, DeepSeek-V3.2's expected regret over the 20-trial post-reversal window fell from 812 coins (fixed) to 460 coins (random), despite an increase in perseveration length. Similar reductions were observed for Gemini-3 and GPT-5.2. Total wins also increased under the random schedule for all LLMs. Thus, greater environmental volatility amplified persistence while simultaneously reducing regret, demonstrating that rigid post-switch behaviour can coexist with improved aggregate performance.

\begin{table}[h]
\setlength{\abovecaptionskip}{0pt}
\setlength{\belowcaptionskip}{0pt}
\setlength{\intextsep}{0pt}
\centering
\caption{Core behavioural measures by agent and schedule. Post-reversal regret is computed from known reward probabilities over a 20-trial window following each true state switch and is reported in expected coins (higher = worse).}
\label{tab:beh_metrics}
\small
\begin{tabular}{|l|l|c|c|c|c|}
\hline
\textbf{Agent} & \textbf{Schedule} & \textbf{Lose-shift} & \textbf{Perseveration} & \textbf{Post-rev.\ regret} & \textbf{Total wins} \\
\hline
DeepSeek-V3.2 & Fixed  & 0.035 & 19.63 & 811.73 & 84.06 \\
DeepSeek-V3.2 & Random & 0.038 & 27.39 & 459.69 & 117.58 \\
\hline
Gemini-3 & Fixed  & 0.235 & 7.24 & 403.43 & 126.11 \\
Gemini-3 & Random & 0.210 & 10.05 & 316.82 & 136.10 \\
\hline
GPT-5.2 & Fixed  & 0.241 & 7.34 & 459.27 & 119.98 \\
GPT-5.2 & Random & 0.209 & 10.90 & 369.23 & 131.13 \\
\hline
Human & Task 2 & 0.495 & 2.87 & 353.84 & 137.39 \\
\hline
\end{tabular}
\end{table}

Although fixed and random schedules shared identical state spaces and switch-trigger rules, they differed in transition predictability. Random transitions systematically increased perseveration across LLMs while reducing post-reversal regret and increasing total wins. This dissociation indicates that aggregate performance metrics alone are insufficient to diagnose adaptive flexibility.

\subsection{Computational mechanisms underlying rigid adaptation}
\label{sec:results_model}

\begin{figure}[h!]
    \centering
    \includegraphics[width=\linewidth]{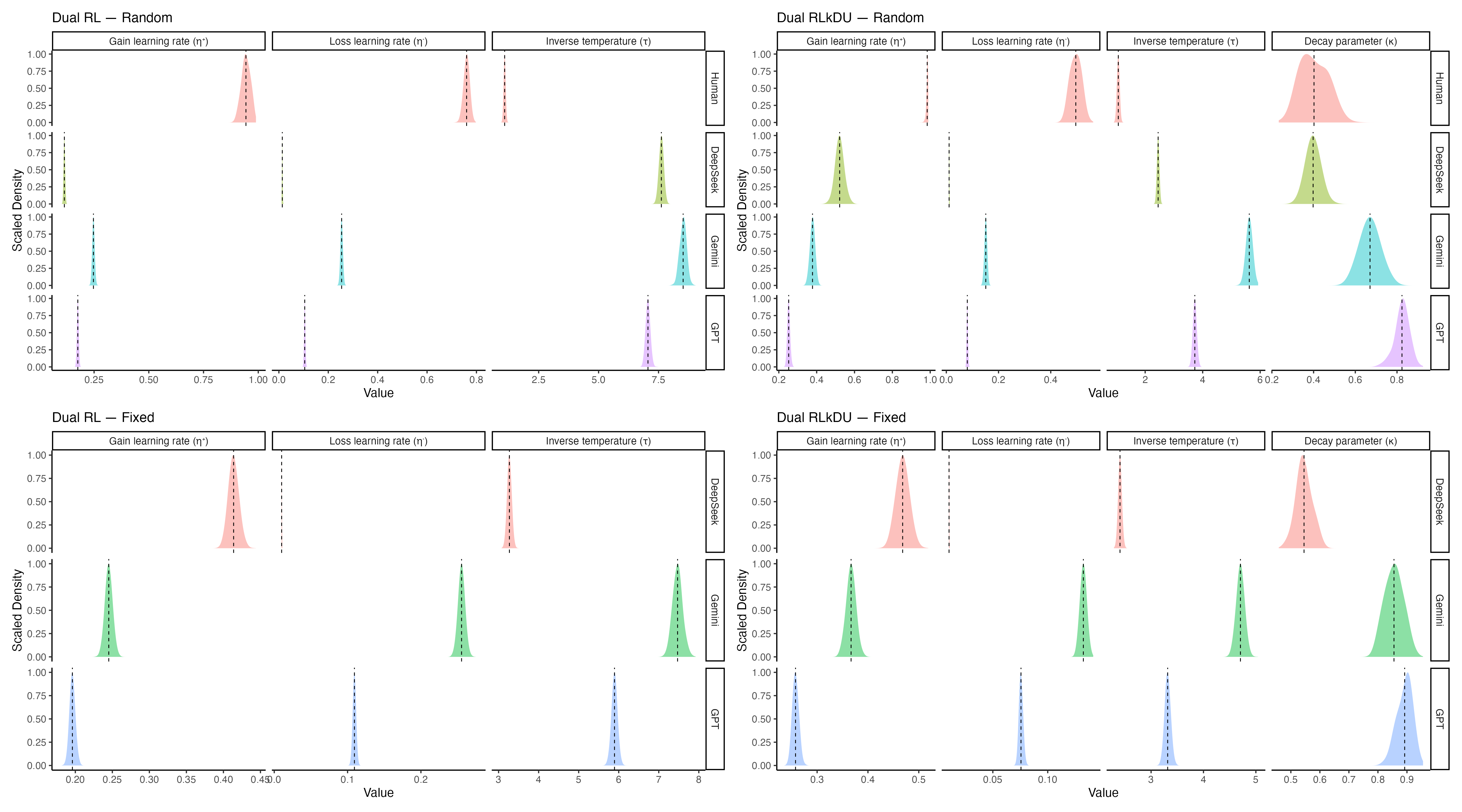}
    \caption{
    Posterior group-level parameter densities under Dual RL and Dual RL-$\kappa$DU (left vs. right rows) for fixed and random schedules (bottom vs.\ top rows). Panels show gain and loss learning rates ($\eta_{\mathrm{pos}}, \eta_{\mathrm{neg}}$), inverse temperature ($\beta$), and (for $\kappa$DU) the counterfactual updating parameter ($\kappa$).
    }
    \label{fig:rl_posteriors}
\end{figure}

To identify the computational origins of behavioural rigidity, we fitted two hierarchical RL models to trial-by-trial choices: Dual RL and Dual RL-$\kappa$DU (Table~\ref{tab:rl_params}). Figure~\ref{fig:rl_posteriors} visualises the corresponding posterior group-level parameter densities across agents and schedule conditions.

\paragraph{Interpretational scope of parameter estimates.}
It is important to clarify that the RL parameters reported here should be interpreted as model-implied process descriptors rather than literal psychological constants. In probabilistic reversal tasks, stationary Rescorla–Wagner models are known to partially conflate incremental value updating with latent state inference and volatility sensitivity. In such contexts, very high gain learning rates do not imply that agents rely exclusively on the most recent outcome. Rather, they indicate that the model approximates rapid belief revision, potentially reflecting implicit change detection, through an elevated updating coefficient. Accordingly, parameters such as $\eta$, $\beta$, and $\kappa$ may absorb aspects of structural changes not explicitly represented in the model. The present interpretation therefore focuses on relative parameter differences across agents and schedules within a shared modelling framework, rather than treating individual parameter magnitudes as direct psychological measurements.

\paragraph{Dual RL: asymmetry and policy determinism.}

Under the baseline Dual RL model, all LLMs exhibited elevated inverse temperature estimates, indicating highly deterministic policies conditional on learned values (Fig.~\ref{fig:rl_posteriors}, left; Table~\ref{tab:rl_params}).
DeepSeek-V3.2 showed moderate determinism in the fixed schedule, which increased markedly under the random schedule.
Gemini-3 and GPT-5.2 were highly deterministic in the fixed schedule, with further increases under randomness.

\begin{table}[h!]
\setlength{\abovecaptionskip}{0pt}
\setlength{\belowcaptionskip}{0pt}
\setlength{\intextsep}{0pt}
\centering
\caption{Hierarchical posterior group-level estimates of computational parameters (mean (SD)).}
\label{tab:rl_params}
\small
\begin{tabular}{|l|l|c|c|c|c|}
\hline
\textbf{Agent} & \textbf{Schedule} & $\eta_{\mathrm{pos}}$ & $\eta_{\mathrm{neg}}$ & $\beta$ & $\kappa$ \\
\hline
\multicolumn{6}{|c|}{\textbf{Dual RL}} \\
\hline
DeepSeek-V3.2 & Fixed  
& 0.421 (0.097) & 0.026 (0.030) & 3.342 (0.769) & -- \\
DeepSeek-V3.2  & Random 
& 0.143 (0.092) & 0.030 (0.022) & 7.519 (0.696) & -- \\
Gemini-3 & Fixed  
& 0.250 (0.016) & 0.260 (0.020) & 7.421 (0.070) & -- \\
Gemini-3 & Random 
& 0.252 (0.017) & 0.259 (0.022) & 8.456 (0.041) & -- \\
GPT-5.2 & Fixed  
& 0.202 (0.014) & 0.117 (0.012) & 5.876 (0.201) & -- \\
GPT-5.2 & Random 
& 0.182 (0.022) & 0.112 (0.015) & 7.017 (0.154) & -- \\
Human & Task 2
& 0.935 (0.002) & 0.755 (0.014) & 1.326 (0.929) & -- \\
\hline
\multicolumn{6}{|c|}{\textbf{Dual RL-$\kappa$DU}} \\
\hline
DeepSeek-V3.2 & Fixed  
& 0.469 (0.025) & 0.017 (0.003) & 2.458 (0.198) & 0.545 (0.011) \\
DeepSeek-V3.2 & Random 
& 0.521 (0.018) & 0.020 (0.003) & 2.502 (0.193) & 0.400 (0.012) \\
Gemini-3 & Fixed  
& 0.369 (0.011) & 0.139 (0.008) & 4.711 (0.137) & 0.848 (0.002) \\
Gemini-3 & Random  
& 0.381 (0.007) & 0.154 (0.007) & 5.556 (0.100) & 0.714 (0.007) \\
GPT-5.2 & Fixed  
& 0.262 (0.010) & 0.083 (0.006) & 3.349 (0.130) & 0.884 (0.001) \\
GPT-5.2 & Random 
& 0.257 (0.011) & 0.088 (0.007) & 3.752 (0.127) & 0.817 (0.002) \\
Human & Task 2  
& 0.975 (0.001) & 0.498 (0.074) & 1.296 (0.594) & 0.404 (0.018) \\
\hline
\end{tabular}
\end{table}

At the level of value updating, the models differed sharply.
DeepSeek-V3.2 exhibited pronounced gain-loss asymmetry ($\eta_{\mathrm{pos}} \gg \eta_{\mathrm{neg}}$), with near-zero loss learning across both schedules as Gemini-3 displayed approximately symmetric learning rates ($\eta_{\mathrm{pos}} \approx \eta_{\mathrm{neg}}$).
This parameter pattern provides a mechanistic account of behavioural rigidity: low lose-shift is expected when $\eta_{\mathrm{neg}}$ is small (losses induce minimal value revision) and/or when $\beta$ is large (strong policy commitment to the currently favoured option).

\paragraph{Dual RL-$\kappa$DU: structural updating reshapes rigidity.}

Allowing counterfactual updating via $\kappa$ altered the inferred parameter structure in systematic ways.

First, inverse temperature estimates decreased across LLMs relative to Dual RL (Fig.~\ref{fig:rl_posteriors}, right; Table~\ref{tab:rl_params}). This indicates that a portion of the apparent policy determinism in Dual RL reflects missing structural changes that are captured when counterfactual updating is permitted. Second, $\kappa$ estimates differentiated agents strongly. GPT-5.2 and Gemini-3 exhibited very high $\kappa$ values in the fixed schedule, consistent with substantial counterfactual updating of the unchosen option.
DeepSeek-V3.2 showed a more moderate $\kappa$.
Where both schedules were estimated, $\kappa$ decreased under the random schedule, suggesting reduced reliance on counterfactual suppression under higher volatility. Humans exhibited comparatively low $\kappa$ and low $\beta$, alongside very high gain learning and substantial loss learning. This profile supports rapid post-loss switching and minimal perseveration, consistent with the behavioural benchmarks.

\paragraph{Mechanistic dissociation of rigidity.} The posterior structure implies distinct computational regimes among LLMs. DeepSeek-V3.2 combines extremely weak loss learning with schedule-sensitive increases in policy determinism under Dual RL, consistent with persistent lose-insensitivity and prolonged perseveration. Gemini-3 shows non-trivial loss learning but pairs it with very high policy determinism under Dual RL and high counterfactual amplification under $\kappa$DU, indicating rigidity driven more by amplified value separation than by loss insensitivity per se. GPT-5.2 exhibits an intermediate profile, with moderate gain-loss asymmetry and very high $\kappa$, consistent with persistence that is supported by structural updating as well as elevated determinism.

\paragraph{Volatility effects.} Under Dual RL, the random schedule increased $\beta$ across all LLMs, consistent with the behavioural increase in perseveration under random reversals. Under Dual RL-$\kappa$DU, this schedule effect was attenuated: rather than requiring a large inflation in $\beta$, the model accommodated volatility through changes in $\kappa$ and smaller shifts in $\beta$, indicating that counterfactual updating stabilises policy expression under unpredictability.

\section{Discussion}
\label{sec:discussion}

While this minimal task does not capture the full complexity of real non-stationary environments, it isolates how agents respond to contingency change under controlled volatility. We show that similar rigid behaviour can arise from distinct computational mechanisms. This distinction has implications for generalisation. In adversarial settings, failure may reflect underweighting of negative evidence, whereas in domains such as market shifts, rigidity may arise from over-commitment or value polarisation despite accumulating signals. In richer environments, these mechanisms may interact with additional structure, such as context or delayed feedback, shaping when and how adaptive behaviour breaks down.

\subsection{Rigidity reflects uncertainty miscalibration} 
A simple explanation for rigid reversal behaviour is slow learning \cite{meister2022learning}. On that view, agents fail to adapt because value updates are too small. Our modelling suggests a deeper interpretation. The critical issue is how the agent decides whether recent negative evidence is informative about a change in the environment or whether it should be treated as noise.

In a reversal task, negative outcomes have two possible meanings. They can be expected stochasticity under the current state, or they can be evidence that the state has changed \cite{zika2023trait}. A flexible agent needs a control rule that decides which meaning to adopt. Humans often behave as if they track this distinction, treating surprising losses as a stronger signal when they occur at the wrong time or in the wrong pattern. In contrast, the LLMs appear to shift toward stronger commitment under unpredictability. This is not well explained by learning rates alone. It points to miscalibrated uncertainty control. Volatility is treated as a reason to stabilise behaviour, rather than to increase hypothesis testing through more exploratory choice. 

One plausible reason is that these LLMs do not maintain an explicit belief state about the latent contingency. They may rely on a compressed summary of recent outcomes that is sufficient for stable environments but insufficient for detecting change points. Another possibility is that the decision protocol encourages coherence and persistence. LLMs are trained to produce consistent outputs given context, and consistency can become a default solution when the environment becomes hard to predict. In a reversal-learning setting, that tendency can look like rigidity.

\subsection{Counterfactual updating as a mechanism for value polarisation}
Adding counterfactual updating changes what it means for an agent to be persistent. Under Dual RL, consistent choice is often explained as a strong mapping from value differences to actions. Under Dual RL-$\kappa$DU, consistent choice can also arise because the unchosen option is pushed away in the opposite direction. This creates value polarisation. Once an option is preferred, the alternative becomes actively less attractive even if direct evidence against it is limited.

This has an important conceptual implication. Some agents may be rigid because they maintain commitment by suppressing alternatives. That mechanism resembles a computational form of confirmation bias. It supports stability when the world is stable, but it becomes costly when the world changes. In practical terms, a high-$\kappa$ agent can look decisive and efficient until it needs to revise its beliefs. At that point, polarised values slow reorientation because the agent must undo both the old preference and the structural suppression of the alternative.

The schedule dependence of $\kappa$ also invites interpretation. A reduction of counterfactual suppression under randomness can be read as a partial attempt to avoid over-polarising values when outcomes are less diagnostic. That is a reasonable adaptation, but it does not guarantee flexibility if the policy still translates value differences into highly committed choices.

\subsection{Determinism is not a single property of the agent}
\label{sec:disc_determinism}

Fitting both model families shows why it is risky to interpret high inverse temperature as a direct measure of stubbornness. Inverse temperature is a modelling construct that absorbs whatever produces consistent choices after accounting for value learning. When counterfactual parameters are omitted, inverse temperature can become a catch-all for missing structure. When counterfactual parameters are included, some of that explanatory load can shift away from inverse temperature.

This has a methodological consequence. Claims about determinism should be framed as claims about a model class. Determinism can reflect a genuinely sharp policy, structural value polarisation, or unmodelled action repetition biases that are not value-based. Stronger mechanistic conclusions require model comparisons that test these alternatives directly, rather than treating a single parameter as a definitive psychological label.

\section{Conclusion}
\label{sec:disc_design}

The computational decomposition suggests that there is no universal fix for rigidity. The right intervention depends on the dominant mechanism. If rigidity is driven by weak use of negative evidence, then the agent needs stronger error attribution. We need prompting or training that forces the agent to explain what a loss implies about the current state. If rigidity is driven by counterfactual suppression, then the agent needs safeguards that keep alternatives viable. This becomes an explicit evaluation of the unchosen option on each trial.

For evaluation, these results argue for diagnostics that go beyond total reward. A strong agent should change its policy for the right reasons and at the right time. Mechanistic fits can support this by distinguishing loss sensitivity and structural suppression. This has implications for benchmarking: standard performance metrics may mask rigid decision processes, allowing high aggregate reward to coexist with poor adaptability. Reversal-sensitive and mechanism-aware benchmarks can therefore provide a more reliable assessment of decision quality, particularly as model capacity increases. Higher-capacity models may achieve strong performance while relying on different underlying strategies. We should evaluate not only outcomes but also the processes that generate them.

Future work should extend these diagnostics to richer environments and incorporate them into benchmark design, including multi-option tasks and variations in prompt formulation.

%
%
%
\bibliographystyle{splncs04}
\bibliography{mybib}
\end{document}